\documentclass[conference]{IEEEtran}
\IEEEoverridecommandlockouts
\usepackage{graphicx}
\usepackage{amsmath}
\usepackage{amssymb}
\usepackage{makeidx}
\usepackage{subfig}
\usepackage{float}
\usepackage{tabularx}
\usepackage{makecell}
\usepackage{multirow}
\usepackage{algorithm}
\usepackage{algpseudocode}
\usepackage[justification=centering]{caption}
\usepackage{nth}
\usepackage{balance}
\def\BibTeX{{\rm B\kern-.05em{\sc i\kern-.025em b}\kern-.08em
    T\kern-.1667em\lower.7ex\hbox{E}\kern-.125emX}}
\begin{document}

\title{The Past and the Present of the Color Checker Dataset Misuse
\thanks{This work has been supported by the Croatian Science Foundation under Project IP-06-2016-2092.}
}

\author{\IEEEauthorblockN{Nikola Bani{\'{c}}, Karlo Ko{\v{s}}{\v{c}}evi{\'{c}}, Marko Suba{\v{s}}i{\'{c}}, and Sven Lon{\v{c}}ari{\'{c}}}
\IEEEauthorblockA{Image Processing Group\\
Faculty of Electrical Engineering and Computing\\
University of Zagreb, 10000 Zagreb, Croatia\\
E-mail: \{nikola.banic, karlo.koscevic, marko.subasic, sven.loncaric\}@fer.hr}
}

\maketitle

\begin{abstract}
The pipelines of digital cameras contain a part for computational color constancy, which aims to remove the influence of the illumination on the scene colors. One of the best known and most widely used benchmark datasets for this problem is the Color Checker dataset. However, due to the improper handling of the black level in its images, this dataset has been widely misused and while some recent publications tried to alleviate the problem, they nevertheless erred and created additional wrong data. This paper gives a history of the Color Checker dataset usage, it describes the origins and reasons for its misuses, and it explains the old and new mistakes introduced in the most recent publications that tried to handle the issue. This should, hopefully, help to prevent similar future misuses.
\end{abstract}


\section{Introduction}
The human vision system (HSV) has the ability to perceive the object color regardless of the scene illumination. This ability is called color constancy~\cite{ebner2007color}. To achieve computational color constancy most contemporary digital cameras estimate scene illumination followed by chromatic adaptation. For this, the image formation is defined as the combination of three physical variables i.e. spectral reflectance properties of the surfaces in the scene, the spectral properties of the light source, and spectral sensitivity of the camera sensor. By using the Lambertian assumption, their relationship is defined as
\begin{equation}
\label{eq:image}
    f_c(\mathbf{x}) = \int_\omega I(\lambda, \mathbf{x})R(\mathbf{x}, \lambda) \rho_c(\lambda)d\lambda
\end{equation}
where the value for color channel $c$ at pixel location $\mathbf{x}$ is the combination of the spectral distribution of the light source $I(\lambda, \mathbf{x})$, the surface reflectance $R(\mathbf{x}, \lambda)$, and the camera sensitivity of the c-th color channel $\rho_c(\lambda)$ across all wavelengths of the light $\lambda$ in the visible spectrum $\omega$. As seen in Eq.~\eqref{eq:image}, different light sources can cause varying color casts in the image. To simplify the problem of illumination estimation, it is often assumed that the illumination is uniform across the whole scene, which gives the observed light color source as
\begin{equation}
\label{eq:e}
    \mathbf{e} = \begin{pmatrix} e_R \\ e_G \\ e_B\end{pmatrix} = \int_\omega I(\lambda)\mathbf{\rho}(\lambda)d\lambda.
\end{equation}
Unfortunately, as both both $I(\lambda)$ and $\mathbf{\rho}(\lambda{})$ are unknown and only image pixel values $\mathbf{f}$ are given, calculating $\mathbf{e}$ is an ill-posed problem. Consequently, many methods which introduce additional assumptions to the problem of illumination estimation have been proposed. They can be divided in two groups, namely the statistics-based and learning-based methods. 

Statistics-based methods include White-patch~\cite{land1977retinex,funt2010rehabilitation,banic2013using,banic2014color,banic2014improving}, Gray-world~\cite{buchsbaum1980spatial}, Shades-of-Gray~\cite{finlayson2004shades}, Gray-Edge~\cite{van2007edge}, using bright and dark colors~\cite{cheng2014illuminant}, exploiting illumination color statistics perception~\cite{banic2019blue}, using gray pixels~\cite{quian2019revisiting}. While statistics-based methods are much faster, computationally less demanding, and hardware-friendly, it are the learning-based methods that achieve much higher accuracies. Such methods include gamut mapping~(pixel, edge, and intersection based)~\cite{finlayson2006gamut}, using high-level visual information~\cite{van2007using}, natural image statistics~\cite{gijsenij2007color}, Bayesian learning~\cite{gehler2008bayesian}, spatio-spectral learning~(maximum likelihood estimate, and with gen. prior)~\cite{chakrabarti2012color}, simplifying the illumination solution space~\cite{banic2015color,banic2015using,banic2015acolor}, using color/edge moments~\cite{finlayson2013corrected}, regression trees with simple features from color distribution statistics~\cite{cheng2015effective}, spatial localizations~\cite{barron2015convolutional,barron2017fast}, genetic algorithms~\cite{koscevic2019color}, convolutional neural networks~\cite{bianco2015color,shi2016deep,hu2017fc4,qiu2018pilot}.

To test the accuracy of various methods in order to compare them, several benchmark datasets have been proposed with one of the best known and most widely used being the Color Checker dataset~\cite{gehler2008bayesian}. However, at one point the black level in the dataset's images started to be handled improperly thus opening the way for wrongly calculated results, misleading comparisons of various methods, several versions of the dataset in circulation, and severe pressure from uninformed paper reviewers. While a recent publication tried to solve the introduced problems, when analyzed carefully it can be shown to introduce even further erroneous data related to the Color Checker dataset. The goal of this paper is to give a short history of how this came to be, to explain how the recent publications do not help to solve the Color Checker problem and make new mistakes, and to describe some additional problems that occur when the Color Checker dataset is used.

The paper is structured as follows: Section~\ref{sec:black} describes the basics of the black level and the need for its subtraction, in Section~\ref{sec:history} a short history of the Color Checker dataset usage is given, Section~\ref{sec:problems} shows the present problems with the Color Checker dataset, and, finally, Section~\ref{sec:conclusions} concludes the paper.

\begin{figure*}[htb]
    \centering
    
  \subfloat[]{
  \includegraphics[width=0.2\linewidth]{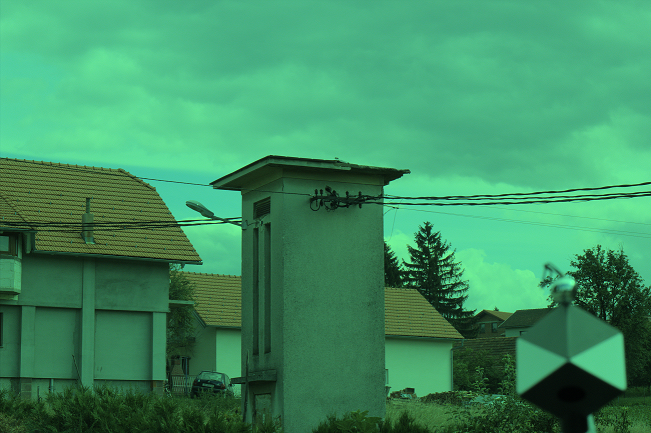}
  \label{fig:nbl_n}
  }%
  \subfloat[]{
  \includegraphics[width=0.2\linewidth]{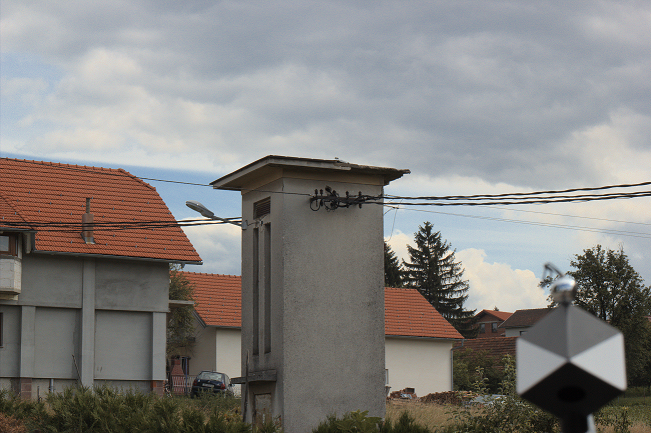}
  \label{fig:nbl_gw}
  }%
  \subfloat[]{
  \includegraphics[width=0.2\linewidth]{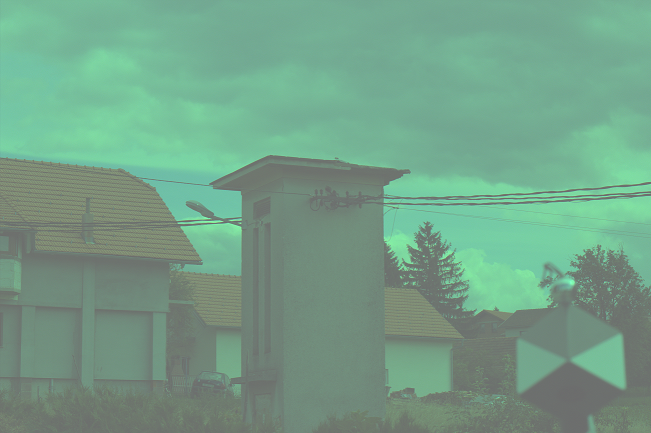}
  \label{fig:bl_n}
  }%
  \subfloat[]{
  \includegraphics[width=0.2\linewidth]{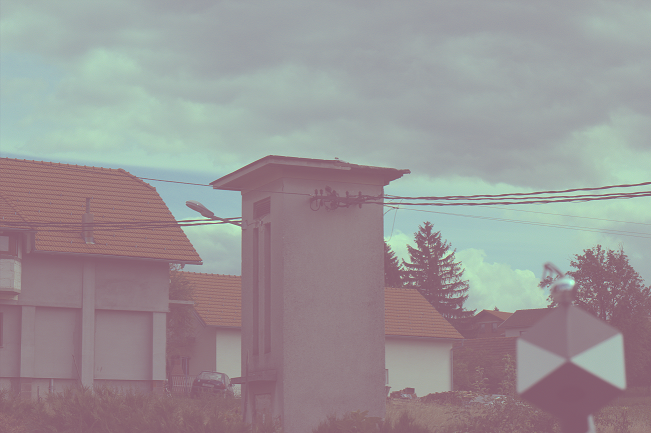}
  \label{fig:bl_gw}
  }
  
    \caption{Effects of (no) black level subtraction: (a)~black level subtraction and no color correction, (b)~black level subtraction and color correction, (c)~no black level subtraction and no color correction, and (d)~no black level subtraction and color correction. For display purposes all images have been tone mapped by applying the Flash tone mapping operator~\cite{banic2016puma, banic2018flash}.}
  \label{fig:bl}
    
\end{figure*}

\section{The black level}
\label{sec:black}

One of the preprocessing techniques applied to the analog signal from the camera sensors is the generation of the black level i.e. the true value of the zero intensity. This is usually done by averaging the dark current signal from the optical black or shaded pixels on the sensor~\cite{allen2012manual}. During the digital image processing in the camera this black level is then subtracted from the image~\cite{terzis2016handbook}. If an image is recorded in its raw form, the black level can later easily be read in the Exif metadata of the image file. While the black level may change from image to image depending on the sensor temperature, it rarely differs significantly from the average black level that can also be found in the Exif data in various appropriately named fields. For these reasons in color constancy benchmark datasets usually a single black level value per a given camera is used on all its images and their channels.

Most of the image processing and computer vision algorithms use images that have already gone through the whole image processing pipeline of a camera. In that case there is no need to perform any kind of black level subtraction since it has already been done in the image processing pipeline. However, some methods like the ones for computational color constancy are applied to linear images i.e. images with no non-linear processing applied at the very beginning of the image processing pipeline. In accordance with Eq.~\eqref{eq:image} these methods expect the black level to be zero and therefore before applying them, the black level has to be subtracted. Failure to do so results in the images looking hazy and it hampers the color correction process as shown in Fig.~\ref{fig:bl} since the images are not linear any more due the presence of the black level.

\begin{figure*}[htb]
    \centering
    
    \includegraphics[width=0.72\linewidth]{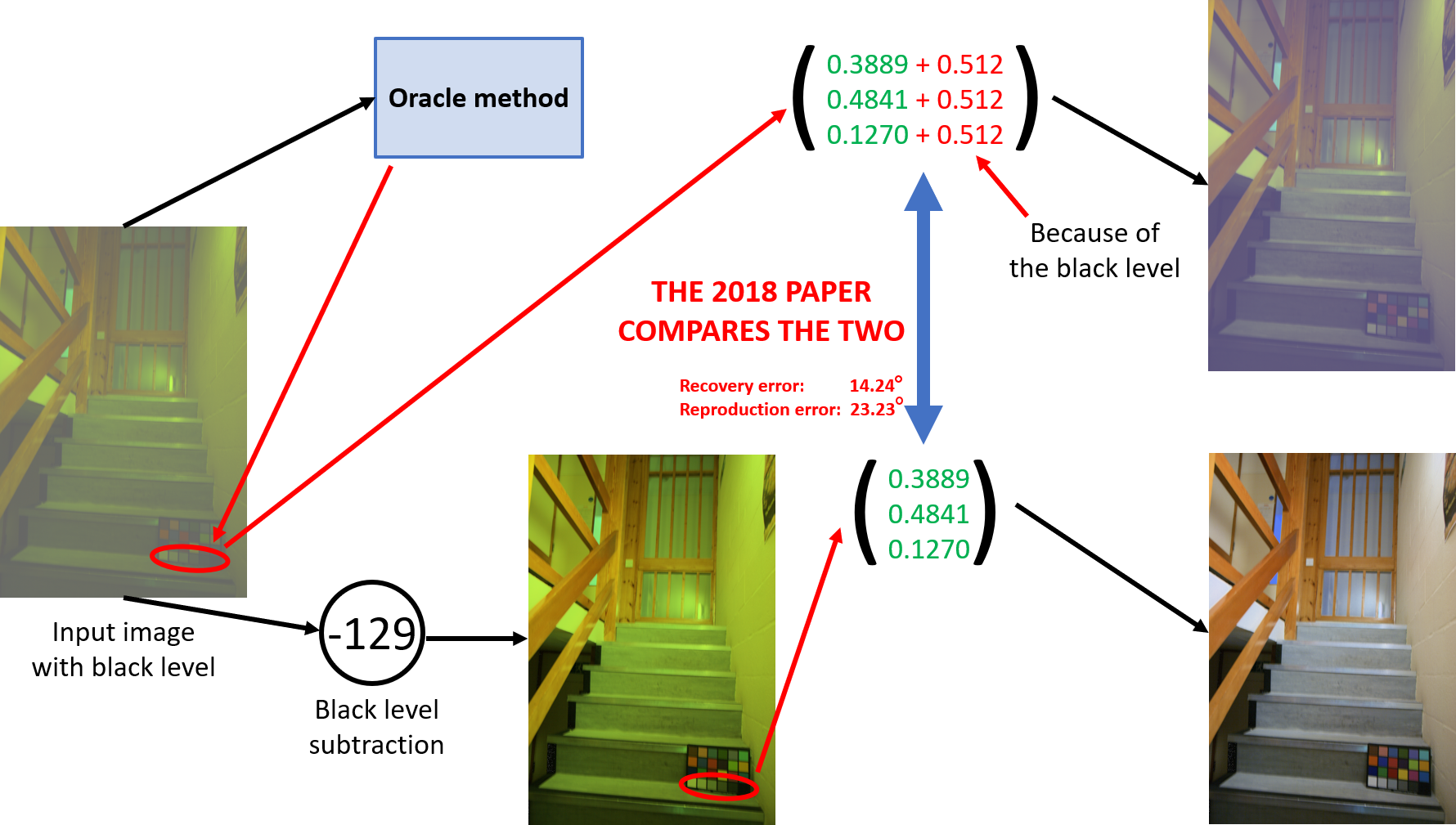}
    \caption{Example of the wrong procedure for error calculation performed in the 2018 paper~\cite{hemrit2018rehabilitating} by using the oracle method.}
    \label{fig:oracle}
    
\end{figure*}

\section{A Short history of the Color Checker dataset}
\label{sec:history}


The Color Checker dataset was released by Gehler et al. in 2008~\cite{gehler2008bayesian}. It consists of $568$ both indoor and outdoor images taken with two high quality DSLR cameras with all settings in auto mode. In the original paper the non-linear versions of the images were used. The images were also provided in the TIFF format that contains information about the linear image versions, but these were not used in the paper. For the purpose of ground-truth illumination extraction, in the scene of each image a color checker was placed and its achromatic patches in the lowest row of the chart were used to calculate the scene ground-truth illumination. On the webpage administrated by Arjan Gijsenij~\cite{gijsenij2019online} the illumination estimations of various methods obtained for this version of the Color Checker dataset and the respective angular error statistics are given under the label ''Color-checker (Original)''.

To make the images of the Color Checker dataset comply with Eq.~\eqref{eq:image}, in 2011 the reprocessed version of the Color Checker dataset was published by Shi and Funt~\cite{shi2018online} by creating linear image versions based on the data in the TIFF format. The main problem was that the remark about the need to subtract the black level was not recognized or noticed by the larger research community and therefore the majority of the experimental results were calculated on images without subtracted black level, which is incorrect. At~\cite{gijsenij2019online} the results for this version are under the label ''Color-checker (by Shi)''.

In 2013 Lynch et at.~\cite{lynch2013colour} were among the first ones to correctly notice and explicitly mention the problem of missing black level subtraction. In their paper they correctly subtracted the black level, then they again extracted the correct ground-truth, and finally they calculated the results of their method and recalculated the results of some other methods on the subset of the correctly processed Color Checker dataset. Unfortunately, their example has not been widely followed in later papers.

In 2014 Cheng et at. published nine new datasets taken by nine different cameras~\cite{cheng2014illuminant}. While calculating the results for their newly processed method as well as the results for earlier methods on the new datasets, they correctly subtracted the black level thus starting a good trend. When they calculated the results of their newly proposed method on the Color Checker dataset, they also did it on the same correct version as Lynch et al. However, for other methods they simply copied the results obtained on the wrong version without the subtracted black level, which therefore further remained in usage and which also gave the results of their method a clear advantage.

In 2017 a paper co-authored by Gehler acknowledged the circulation of at least three different ground-truths for the Color Checker dataset~\cite{finlayson2017curious}. One of the problems noticed there is that the ranking of illumination estimation methods varies depending on the ground-truth that is used. This opened the question whether there were real advances in  illumination estimation or did the authors just use the ground-truth on which their methods achieve better performances~\cite{finlayson2017curious}.

In 2018 a paper authored by several leading experts in the area of illumination estimation has been published~\cite{hemrit2018rehabilitating} with the goal of proposing a correctly calculated ground-truth for the Color Checker dataset. This ground-truth is then used to review the performance evaluation of a range of illumination estimation methods and it is concluded that the difference in how methods perform can be large with many local rankings being reversed with respect to the ranking obtained for the older versions of the ground-truth. During the review process of new papers at least some reviewers take this paper for granted and claim that together with its methods ranking and new ground-truth it makes the Color Checker dataset again valid and necessary to be used in future experimental results thus exercising pressure if this is not done. However, the test procedure for some of the allegedly tested methods in the 2018 paper and consequently the obtained accuracies are wrong.

\section{Present problems}
\label{sec:problems}

\subsection{Wrong experimental data}
\label{subsec:calculation}

While the newly proposed ground-truth illuminations for the Color Checker dataset are necessary, they are not novel. If the ground-truth proposed by the project related to the paper of Lynch et at.~\cite{lynch2013colour} that could have been downloaded earlier is compared to the newly recommended ground-truth, the median of the per image angular difference between the two versions is $0.04^\circ$ with the $25$\% highest difference being $0.21^\circ$, which means that these two are very similar and effectively the same.

Next, while the recommended ground-truth is indeed correct and should be used, the allegedly obtained illumination estimation accuracy results for some of the methods are wrong. The illumination estimations that were used to calculate the results mentioned in the 2018 paper can be downloaded at~\cite{gijsenij2019online}. These illumination estimations were supposed to be obtained by running the methods on the images with the black level subtracted because the newly recommended ground-truth was also obtained in that way. However, for practically all methods these illumination estimations are the same as the ones obtained when the black level was not subtracted and the illumination estimations were compared to the previously used wrong ground-truth.

This effectively means that the experimental results mentioned in the 2018 paper were obtained as if the methods were run against the images that did not have the black level subtracted and then their illumination estimations were compared to the ground-truth that was obtained after the black level has been subtracted. Even without downloading the mentioned data and making the comparison, just by looking at the supposed results of e.g. the Gray-world method that are significantly worse than on other datasets with linear images, it can be concluded that something may have gone wrong.

The Gray-World method allegedly scores a $9.97^\circ$ median angular error even though the dataset contains raw images that were not processed in any non-linear way. In an experiment we tried to calculate the results for the Gray-world method on the images when the black level is not subtracted and then compare it to the recommended ground-truth, which is a deliberately chosen wrong procedure, and the obtained median error was $9.94^\circ$, similar to $9.97^\circ$ reported in the 2018 paper. Next, we performed the experiment correctly by first subtracting the black level, then calculating the Gray-world illumination estimations, and only then comparing them to the newly recommended ground-truth. The obtained median angular error was $3.54^\circ$, which seemed more realistic. As a matter of fact, by looking at the Gray-world results on the archived version of the project webpage associated with the 2013 paper~\cite{lynch2015online}, the reported median angular error is $3.98^\circ$, which is relatively similar to the $3.54^\circ$ obtained here.

If the recommended ground-truth from the 2018 paper is compared to the previously used ground-truth that was calculated on the images without the black level being subtracted, the median angular per image difference is $3.17^\circ$ and the maximum difference is $18.21^\circ$. Imagine a perfect oracle method that produces no error whatsoever being applied to the same images that other methods in the 2018 paper were effectively applied to, namely to the images without the black level being subtracted. It gives the perfect estimation, but then it is compared to a ground-truth that has been calculated on the other images, namely the ones with the black level subtracted. As shown, despite the method being a perfect oracle, it is possible that the alleged error will be over $18^\circ$ just because the whole used methodology is wrong. A simplified diagram of such a wrong error calculation is shown in Fig.~\ref{fig:oracle}.

This makes these experimental results wrong by definition and they should definitely not be used for comparison with the results of other newly proposed methods' performance on the correct version of the Color Checker dataset. Since the 2018 paper does not state this, it can be argued that alongside erroneous experimental data it also introduces additional confusion that can be dangerous for the wider community. Namely, the authority of the authors may make the use of these results to be considered mandatory when writing new papers on illumination estimation methods despite them being wrong and there have been cases where some reviewers actually insisted on such actions.

\subsection{Multiple sensors}
\label{subsec:sensors}

\begin{figure}[htb]
    \centering
    
    \includegraphics[width=0.75\linewidth]{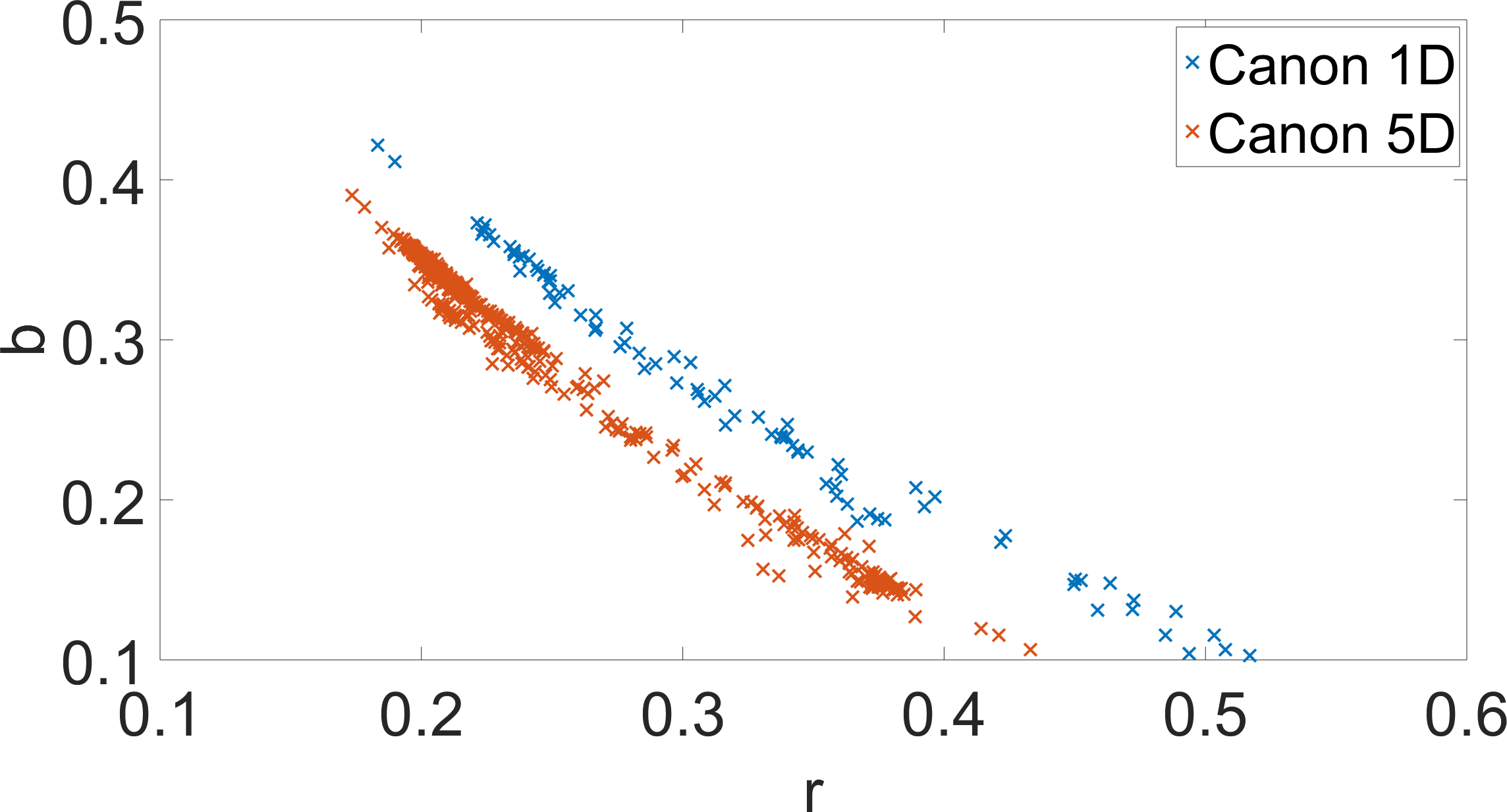}
    \caption{Ground-truth illuminations of the Color Checker dataset in the $rb$-chromaticity plane by used camera models.}
    \label{fig:canons}
    
\end{figure}

Another problem with the Color Checker dataset is that it was created with two different camera models. For the first $86$ images the Canon 1D camera was used, while the rest $482$ images were taken by using the Canon 5D camera. The direct effect of this on the newly recommended ground-truth~\cite{hemrit2018rehabilitating} is visible in Fig.~\ref{fig:canons} with the ground-truths concentrated around two lines, each for one of the cameras. If a method assumes the black body radiation for the illumination model, which is true for many real-world illuminations, in some cases this method also indirectly assumes modelling the illuminations by using a line in the chromaticity plane. By processing the Color Checker data, such a method would fail even though the majority of the illuminations in the Color Checker can be modelled as black body radiation. The reason for this failure is the variability of $\mathbf{\rho}(\lambda)$ in Eq.~\eqref{eq:image} since two cameras were used to take the images. While it may be argued that due to the ill-posedness of the illumination estimation problem this is not a serious issue, it must be stressed that there are no similar issues with other widely used benchmark datasets. As a matter of fact, the case when there are multiple camera sensors involved is treated in the literature as inter-camera color constancy~\cite{gao2017improving,banic2017unsupervised}. Therefore, before using the Color Checker dataset to evaluate a method's performance, it should be considered whether it could be affected by learning from images taken from multiple camera sensors whose ground-truth illumination chromaticity distributions are different.

\subsection{Folds}
\label{subsec:folds}

Another problem with Color Checker is creating data folds. When checking a method's performance on the Color Checker dataset, usually a three-fold cross-validation is used. However, the problem is that not all researchers state in their papers whether they use folds created without image shuffling, with custom shuffling, or with shuffling originally provided by author of the Color Checker dataset and available for download at several places, e.g. at~\cite{barron2019online}. Using different folds can result in different performance metrics values and put the repeatability of the experimental results in question. Additionally, since in case of no shuffling only the first fold has Canon 1D images, the ground-truth illumination distribution differs significantly between the folds, which is statistically not desirable.

\subsection{Violation of the uniform illumination assumption}
\begin{figure}[htb]
    \centering
    
    \includegraphics[width=0.35\linewidth]{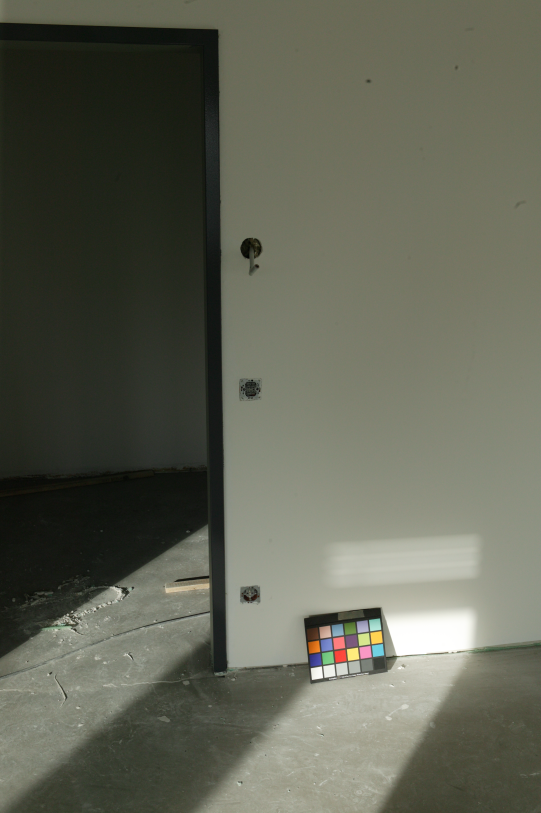}
    \caption{Example of an image in the Color Checker dataset where the assumption of uniform illumination is violated.}
    \label{fig:no_uniform}
    
\end{figure}

Finally, in Color Checker images the uniform illumination assumption does not always hold, which can lead to wrong ground-truth extraction as already mentioned by other authors~\cite{zakizadeh2015hybrid}. For example, if the walls in Fig.~\ref{fig:no_uniform} are assumed to be white, it can be calculated that the angle between the white color of the background wall and the color of the shadowed foreground wall is around $1.8^\circ$. There is also a third illumination that influences the color checker achromatic fields so that the angle between their and the color of the foreground and background white wall is over $2^\circ$ and $4^\circ$, respectively.

\section{Conclusions}
\label{sec:conclusions}

A short history of the various Color Checker dataset (mis)uses has been given and the technical reasons for their emergence have been explained. It has been shown why the recent paper that supposedly solves many of the Color Checker issues only introduces further confusion and erroneous experimental data. Because of that, it should not be considered mandatory to include it in experimental results of newly proposed methods before the results of existing methods are correctly recalculated and published. Additionally, since Color Checker has other problems like involvement of multiple sensors, various folds being used, and the uniform illumination assumption violation in many of its images, it may be better to use newer datasets with fewer similar problems~\cite{cheng2014illuminant,banic2017unsupervised}.

\section*{Acknowledgment}
\label{sec:acknowledgements}

This work has been supported by the Croatian Science Foundation under Project IP-06-2016-2092.

\bibliography{literature}

\begin{thebibliography}{10}
\providecommand{\url}[1]{#1}
\csname url@samestyle\endcsname
\providecommand{\newblock}{\relax}
\providecommand{\bibinfo}[2]{#2}
\providecommand{\BIBentrySTDinterwordspacing}{\spaceskip=0pt\relax}
\providecommand{\BIBentryALTinterwordstretchfactor}{4}
\providecommand{\BIBentryALTinterwordspacing}{\spaceskip=\fontdimen2\font plus
\BIBentryALTinterwordstretchfactor\fontdimen3\font minus
  \fontdimen4\font\relax}
\providecommand{\BIBforeignlanguage}[2]{{%
\expandafter\ifx\csname l@#1\endcsname\relax
\typeout{** WARNING: IEEEtran.bst: No hyphenation pattern has been}%
\typeout{** loaded for the language `#1'. Using the pattern for}%
\typeout{** the default language instead.}%
\else
\language=\csname l@#1\endcsname
\fi
#2}}
\providecommand{\BIBdecl}{\relax}
\BIBdecl

\bibitem{ebner2007color}
M.~Ebner, \emph{Color Constancy}, ser. The Wiley-IS\&T Series in Imaging
  Science and Technology.\hskip 1em plus 0.5em minus 0.4em\relax Wiley, 2007.

\bibitem{land1977retinex}
E.~H. Land, \emph{The retinex theory of color vision}.\hskip 1em plus 0.5em
  minus 0.4em\relax Scientific America., 1977.

\bibitem{funt2010rehabilitation}
B.~Funt and L.~Shi, ``The rehabilitation of {MaxRGB},'' in \emph{Color and
  Imaging Conference}, vol. 2010, no.~1.\hskip 1em plus 0.5em minus 0.4em\relax
  Society for Imaging Science and Technology, 2010, pp. 256--259.

\bibitem{banic2013using}
N.~Bani\'{c} and S.~Lon\v{c}ari\'{c}, ``{U}sing the {R}andom {S}prays {R}etinex
  {A}lgorithm for {G}lobal {I}llumination {E}stimation,'' in \emph{Proceedings
  of The Second Croatian Computer Vision Workshopn (CCVW 2013)}.\hskip 1em plus
  0.5em minus 0.4em\relax University of Zagreb Faculty of Electrical
  Engineering and Computing, 2013, pp. 3--7.

\bibitem{banic2014color}
------, ``{C}olor {R}abbit: {G}uiding the {D}istance of {L}ocal {M}aximums in
  {I}llumination {E}stimation,'' in \emph{Digital Signal Processing (DSP), 2014
  19th International Conference on}.\hskip 1em plus 0.5em minus 0.4em\relax
  IEEE, 2014, pp. 345--350.

\bibitem{banic2014improving}
------, ``Improving the {W}hite patch method by subsampling,'' in \emph{Image
  Processing (ICIP), 2014 21st IEEE International Conference on}.\hskip 1em
  plus 0.5em minus 0.4em\relax IEEE, 2014, pp. 605--609.

\bibitem{buchsbaum1980spatial}
G.~Buchsbaum, ``A spatial processor model for object colour perception,''
  \emph{Journal of The Franklin Institute}, vol. 310, no.~1, pp. 1--26, 1980.

\bibitem{finlayson2004shades}
G.~D. Finlayson and E.~Trezzi, ``Shades of gray and colour constancy,'' in
  \emph{Color and Imaging Conference}, vol. 2004, no.~1.\hskip 1em plus 0.5em
  minus 0.4em\relax Society for Imaging Science and Technology, 2004, pp.
  37--41.

\bibitem{van2007edge}
J.~Van De~Weijer, T.~Gevers, and A.~Gijsenij, ``Edge-based color constancy,''
  \emph{Image Processing, IEEE Transactions on}, vol.~16, no.~9, pp.
  2207--2214, 2007.

\bibitem{cheng2014illuminant}
D.~Cheng, D.~K. Prasad, and M.~S. Brown, ``Illuminant estimation for color
  constancy: why spatial-domain methods work and the role of the color
  distribution,'' \emph{JOSA A}, vol.~31, no.~5, pp. 1049--1058, 2014.

\bibitem{banic2019blue}
N.~Bani\'{c} and S.~Lon\v{c}ari\'{c}, ``{B}lue {S}hift {A}ssumption:
  {I}mproving {I}llumination {E}stimation {A}ccuracy for {S}ingle {I}mage from
  {U}nknown {S}ource,'' in \emph{VISAPP}, 2019, pp. 191--197.

\bibitem{quian2019revisiting}
Y.~Qian, S.~Pertuz, J.~Nikkanen, J.-K. K{\:{a}}m{\:{a}}r{\:{a}}inen, and
  J.~Matas, ``{R}evisiting {G}ray {P}ixel for {S}tatistical {I}llumination
  {E}stimation,'' in \emph{VISAPP}, 2019, pp. 36--46.

\bibitem{finlayson2006gamut}
G.~D. Finlayson, S.~D. Hordley, and I.~Tastl, ``Gamut constrained illuminant
  estimation,'' \emph{International Journal of Computer Vision}, vol.~67,
  no.~1, pp. 93--109, 2006.

\bibitem{van2007using}
J.~Van De~Weijer, C.~Schmid, and J.~Verbeek, ``Using high-level visual
  information for color constancy,'' in \emph{Computer Vision, 2007. ICCV 2007.
  IEEE 11th International Conference on}.\hskip 1em plus 0.5em minus
  0.4em\relax IEEE, 2007, pp. 1--8.

\bibitem{gijsenij2007color}
A.~Gijsenij and T.~Gevers, ``{C}olor {C}onstancy using {N}atural {I}mage
  {S}tatistics.'' in \emph{CVPR}, 2007, pp. 1--8.

\bibitem{gehler2008bayesian}
P.~V. Gehler, C.~Rother, A.~Blake, T.~Minka, and T.~Sharp, ``Bayesian color
  constancy revisited,'' in \emph{Computer Vision and Pattern Recognition,
  2008. CVPR 2008. IEEE Conference on}.\hskip 1em plus 0.5em minus 0.4em\relax
  IEEE, 2008, pp. 1--8.

\bibitem{chakrabarti2012color}
A.~Chakrabarti, K.~Hirakawa, and T.~Zickler, ``Color constancy with
  spatio-spectral statistics,'' \emph{Pattern Analysis and Machine
  Intelligence, IEEE Transactions on}, vol.~34, no.~8, pp. 1509--1519, 2012.

\bibitem{banic2015color}
N.~Bani{\'{c}} and S.~Lon{\v{c}}ari{\'{c}}, ``{C}olor {C}at: {R}emembering
  {C}olors for {I}llumination {E}stimation,'' \emph{Signal Processing Letters,
  IEEE}, vol.~22, no.~6, pp. 651--655, 2015.

\bibitem{banic2015using}
------, ``Using the red chromaticity for illumination estimation,'' in
  \emph{Image and Signal Processing and Analysis (ISPA), 2015 9th International
  Symposium on}.\hskip 1em plus 0.5em minus 0.4em\relax IEEE, 2015, pp.
  131--136.

\bibitem{banic2015acolor}
N.~Bani\'{c} and S.~Lon\v{c}ari\'{c}, ``{C}olor {D}og: {G}uiding the {G}lobal
  {I}llumination {E}stimation to {B}etter {A}ccuracy,'' in \emph{VISAPP}, 2015,
  pp. 129--135.

\bibitem{finlayson2013corrected}
G.~D. Finlayson, ``Corrected-moment illuminant estimation,'' in
  \emph{Proceedings of the IEEE International Conference on Computer Vision},
  2013, pp. 1904--1911.

\bibitem{cheng2015effective}
D.~Cheng, B.~Price, S.~Cohen, and M.~S. Brown, ``Effective learning-based
  illuminant estimation using simple features,'' in \emph{Proceedings of the
  IEEE Conference on Computer Vision and Pattern Recognition}, 2015, pp.
  1000--1008.

\bibitem{barron2015convolutional}
J.~T. Barron, ``{C}onvolutional {C}olor {C}onstancy,'' in \emph{Proceedings of
  the IEEE International Conference on Computer Vision}, 2015, pp. 379--387.

\bibitem{barron2017fast}
J.~T. Barron and Y.-T. Tsai, ``{F}ast {F}ourier {C}olor {C}onstancy,'' in
  \emph{Computer Vision and Pattern Recognition, 2017. CVPR 2017. IEEE Computer
  Society Conference on}, vol.~1.\hskip 1em plus 0.5em minus 0.4em\relax IEEE,
  2017.

\bibitem{koscevic2019color}
K.~Ko{\v{s}}{\v{c}}evi{\'{c}}, N.~Bani\'{c}, and S.~Lon\v{c}ari\'{c}, ``{C}olor
  {B}eaver: {B}ounding {I}llumination {E}stimations for {H}igher {A}ccuracy,''
  in \emph{VISAPP}, 2019, pp. 183--190.

\bibitem{bianco2015color}
S.~Bianco, C.~Cusano, and R.~Schettini, ``{C}olor {C}onstancy {U}sing {CNN}s,''
  in \emph{Proceedings of the IEEE Conference on Computer Vision and Pattern
  Recognition Workshops}, 2015, pp. 81--89.

\bibitem{shi2016deep}
W.~Shi, C.~C. Loy, and X.~Tang, ``{D}eep {S}pecialized {N}etwork for
  {I}lluminant {E}stimation,'' in \emph{European Conference on Computer
  Vision}.\hskip 1em plus 0.5em minus 0.4em\relax Springer, 2016, pp. 371--387.

\bibitem{hu2017fc4}
Y.~Hu, B.~Wang, and S.~Lin, ``{F}ully {C}onvolutional {C}olor {C}onstancy with
  {C}onfidence-weighted {P}ooling,'' in \emph{Computer Vision and Pattern
  Recognition, 2017. CVPR 2017. IEEE Conference on}.\hskip 1em plus 0.5em minus
  0.4em\relax IEEE, 2017, pp. 4085--4094.

\bibitem{qiu2018pilot}
J.~Qiu, H.~Xu, Y.~Ma, and Z.~Ye, ``{PILOT}: {A} {P}ixel {I}ntensity {D}riven
  {I}lluminant {C}olor {E}stimation {F}ramework for {C}olor {C}onstancy,''
  \emph{arXiv preprint arXiv:1806.09248}, 2018.

\bibitem{banic2016puma}
N.~Bani{\'c} and S.~Lon{\v{c}}ari{\'c}, ``{P}uma: {A} high-quality
  retinex-based tone mapping operator,'' in \emph{Signal Processing Conference
  (EUSIPCO), 2016 24th European}.\hskip 1em plus 0.5em minus 0.4em\relax IEEE,
  2016, pp. 943--947.

\bibitem{banic2018flash}
------, ``{F}lash and {S}torm: {F}ast and {H}ighly {P}ractical {T}one {M}apping
  based on {N}aka-{R}ushton {E}quation,'' in \emph{International Conference on
  Computer Vision Theory and Applications}, 2018, pp. 47--53.

\bibitem{allen2012manual}
E.~Allen and S.~Triantaphillidou, \emph{The Manual of Photography and Digital
  Imaging}.\hskip 1em plus 0.5em minus 0.4em\relax Focal Press, 2012.

\bibitem{terzis2016handbook}
A.~Terzis, \emph{Handbook of Camera Monitor Systems: The Automotive
  Mirror-Replacement Technology Based on ISO 16505}.\hskip 1em plus 0.5em minus
  0.4em\relax Springer, 2016, vol.~5.

\bibitem{hemrit2018rehabilitating}
\BIBentryALTinterwordspacing
G.~Hemrit, G.~D. Finlayson, A.~Gijsenij, P.~V. Gehler, S.~Bianco, and M.~S.
  Drew, ``Rehabilitating the color checker dataset for illuminant estimation,''
  \emph{CoRR}, vol. abs/1805.12262, 2018. [Online]. Available:
  \url{http://arxiv.org/abs/1805.12262}
\BIBentrySTDinterwordspacing

\bibitem{gijsenij2019online}
\BIBentryALTinterwordspacing
T.~G. A.~Gijsenij and J.~van~de Weijer. (2019) Color {C}onstancy | {R}esearch
  {W}ebsite on {I}lluminant {E}stimation. [Online]. Available:
  \url{http://colorconstancy.com/}
\BIBentrySTDinterwordspacing

\bibitem{shi2018online}
\BIBentryALTinterwordspacing
L.~Shi and B.~Funt. (2018, Oct) {R}e-processed {V}ersion of the {G}ehler
  {C}olor {C}onstancy {D}ataset of 568 {I}mages. [Online]. Available:
  \url{http://www.cs.sfu.ca/\~colour/data/}
\BIBentrySTDinterwordspacing

\bibitem{lynch2013colour}
S.~E. Lynch, M.~S. Drew, and k.~G.~D. Finlayson, ``{C}olour {C}onstancy from
  {B}oth {S}ides of the {S}hadow {E}dge,'' in \emph{Color and Photometry in
  Computer Vision Workshop at the International Conference on Computer
  Vision}.\hskip 1em plus 0.5em minus 0.4em\relax IEEE, 2013.

\bibitem{finlayson2017curious}
G.~D. Finlayson, G.~Hemrit, A.~Gijsenij, and P.~Gehler, ``{A} {C}urious
  {P}roblem with {U}sing the {C}olour {C}hecker {D}ataset for {I}lluminant
  {E}stimation,'' in \emph{Color and Imaging Conference}, vol. 2017,
  no.~25.\hskip 1em plus 0.5em minus 0.4em\relax Society for Imaging Science
  and Technology, 2017, pp. 64--69.

\bibitem{lynch2015online}
\BIBentryALTinterwordspacing
S.~E. Lynch, M.~S. Drew, and k.~G.~D. Finlayson. (2015) Reprocessed gehler |
  datasets | uea colour group. [Online]. Available:
  \url{http://tinyurl.com/lynch-old}
\BIBentrySTDinterwordspacing

\bibitem{gao2017improving}
S.-B. Gao, M.~Zhang, C.-Y. Li, and Y.-J. Li, ``Improving color constancy by
  discounting the variation of camera spectral sensitivity,'' \emph{JOSA A},
  vol.~34, no.~8, pp. 1448--1462, 2017.

\bibitem{banic2017unsupervised}
N.~Bani{\'{c}} and S.~Lon{\v{c}}ari{\'{c}}, ``{U}nsupervised {L}earning for
  {C}olor {C}onstancy,'' \emph{arXiv preprint arXiv:1712.00436}, 2017.

\bibitem{barron2019online}
\BIBentryALTinterwordspacing
J.~T. Barron. (2019) Gehler's color checker dataset shuffling. [Online].
  Available: \url{http://tinyurl.com/cc-3-folds}
\BIBentrySTDinterwordspacing

\bibitem{zakizadeh2015hybrid}
R.~Zakizadeh, M.~S. Brown, and G.~D. Finlayson, ``{A} {H}ybrid {S}trategy {F}or
  {I}lluminant {E}stimation {T}argeting {H}ard {I}mages,'' in \emph{Proceedings
  of the IEEE International Conference on Computer Vision Workshops}, 2015, pp.
  16--23.

\end{thebibliography}
\bibliographystyle{IEEEtran}
\balance

\end{document}